\documentclass[pdflatex,sn-mathphys]{sn-jnl}
\jyear{2023}%
\theoremstyle{thmstyleone}

\theoremstyle{thmstyletwo}

\theoremstyle{thmstylethree}

\usepackage{amsmath,amssymb,amsfonts}%
\usepackage{mathrsfs}%
\usepackage{algorithm}%
\usepackage{algorithmicx}%
\usepackage{textcomp}
\usepackage{xcolor}
\usepackage{url}
\usepackage{multirow} 
\usepackage{tabularx}
\usepackage{lipsum}
\usepackage{comment}
\usepackage{subcaption}
\usepackage{graphicx}

\usepackage{colortbl}
\usepackage{geometry}

\begin{document}
\title[Series2Vec: Similarity-based Representation Learning]{Series2Vec: Similarity-based Self-supervised Representation Learning for Time Series Classification

}

\author*[1]{\fnm{Navid} \sur{Mohammadi Foumani}}\email{navid.foumani@monash.edu.com}

\author[1]{\fnm{Chang Wei} \sur{Tan}}\email{chang.tan@monash.edu} 
\author[1]{\fnm{Geoffrey I.} \sur{Webb}}\email{geoff.webb@monash.edu}
\author[1]{\fnm{Hamid} \sur{Rezatofighi}}\email{hamid.rezatofighi@monash.edu}
\author[1]{\fnm{Mahsa} \sur{Salehi}}\email{mahsa.salehi@monash.edu}

\affil*[1]{\orgdiv{Department of Data Science and Artificial Intelligence}, \orgname{Monash~University}, \orgaddress{ \city{Melbourne}, \state{VIC}, \country{Australia}}}

\abstract{

We argue that time series analysis is fundamentally different in nature to either vision or natural language processing with respect to the forms of meaningful self-supervised learning tasks that can be defined. Motivated by this insight, we introduce a novel approach called \textit{Series2Vec} for self-supervised representation learning. 

Unlike other self-supervised methods in time series, which carry the risk of positive sample variants being less similar to the anchor sample than series in the negative set, Series2Vec is trained to predict the similarity between two series in both temporal and spectral domains through a self-supervised task. Series2Vec relies primarily on the consistency of the unsupervised similarity step, rather than the intrinsic quality of the similarity measurement, without the need for hand-crafted data augmentation. 
To further enforce the network to learn similar representations for similar time series, we propose a novel approach that applies order-invariant attention to each representation within the batch during training. 
Our evaluation of Series2Vec on nine large real-world datasets, along with the UCR/UEA archive, shows enhanced performance compared to current state-of-the-art self-supervised techniques for time series.
Additionally, our extensive experiments show that Series2Vec performs comparably with fully supervised training and offers high efficiency in datasets with limited-labeled data. Finally, we show that the fusion of Series2Vec with other representation learning models leads to enhanced performance for time series classification. Code and models are open-source at \url{https://github.com/Navidfoumani/Series2Vec.}
}

\keywords{Representation Learning, Similarity-based Self-supervised, Time series Classification}
\maketitle
\vspace{-0.5cm}
\section{Introduction} 
Learning from large time series datasets is important in various fields such as human activity recognition \cite{foumani2023deep}, diagnosis based on electronic health records \cite{rajkomar2018scalable}, and systems monitoring problems \cite{bagnall2018uea}.
These applications can generate hundreds to thousands of time series every day, producing large quantities of data that are critical for the performance of various time series tasks.
However, obtaining labeled data for large time series datasets can be costly and challenging.
Machine learning models trained on large labeled time series datasets tend to produce better performance than models trained on sparsely labeled datasets, small datasets with limited labels or without supervision which produce subpar performance on various time series machine learning tasks \cite{yue2022ts2vec,yang2022unsupervised}.
Therefore, instead of relying on good quality annotations on large datasets, researchers and practitioners are now turning their attention towards self-supervised representation learning for timeseries.


Self-supervised representation learning is a subfield of machine learning that aims to learn representations from data without requiring explicit supervision \cite{goyal2021self}. 
Unlike supervised learning, where models are trained on labeled data, self-supervised learning methods leverage the inherent structure of the data to learn useful representations in an unsupervised manner.
The learned representations can then be used for a variety of downstream tasks such as classification, anomaly detection, and forecasting \cite{foumani2023deep}.

Contrastive learning is an effective and popular self-supervised learning method, originally developed for image analysis \cite{oord2018representation}. 
In contrastive learning, the model learns to differentiate between similar and dissimilar examples. 
These methods have been successfully used to improve performance in a variety of learning tasks such as image classification \cite{oord2018representation}, object detection \cite{chen2020simple,he2020momentum}, and natural language processing \cite{oord2018representation}. 

In spite of the research progress in self-supervised approaches in vision and language, this area is in its infancy for time series~\cite{foumani2023deep}. In this paper, we propose a new approach to self-supervised learning for time series that is inspired by contrastive learning~\cite{oord2018representation}. 
A common yet powerful method for contrastive learning with images is to first create synthetic transformations (augmentation) of an image and then the model learns to contrast the image and its transforms from other images in the training data. 
We believe that this approach works well for images because many learning tasks related to images involve the interpretation of the objects captured in the image. Transformations such as scaling, blurring, and rotation assume that the resulting images will resemble those that would have been generated in the original scenario with changes in camera zoom, stability, focus, or angle.

\begin{figure}
    \centering
    \includegraphics[width=0.65\textwidth]{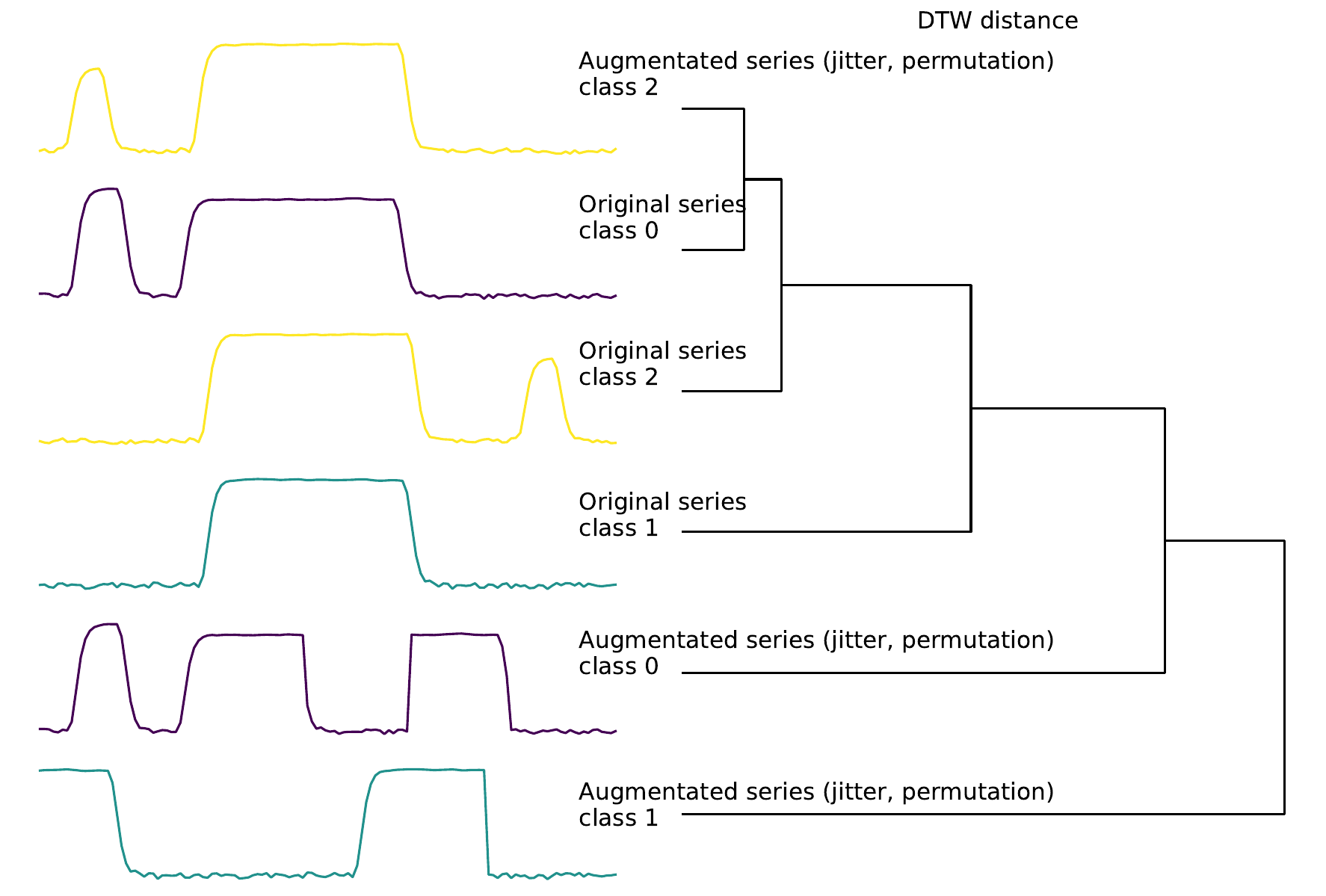}
    \caption{A dendrogram comparing the similarity of three time series of different classes and their augmented variants taken from the BME dataset \cite{dau2019ucr}. The three original raw series are augmented using the strong augmentation technique (jittering and permutation) proposed in TS-TCC \cite{eldele2021time}. 
    Under Dynamic Time Warping distance, the original series of class 0 is most similar to the augmented series of class 2. 
    Additionally, the augmented series of class 0 and 1 are quite dissimilar from their original series.}
    \label{fig:tcc-augmentation}
    \vspace{-0.5cm}
\end{figure}

However, there do not appear to be equivalent transformations that can be applied to time series data. Transformations that have been used in contrastive learning for time series, including TS-TCC \cite{eldele2021time}, MCL~\cite{wickstrom2022mixing}, TS2Vec \cite{yue2022ts2vec}, BTSF \cite{yang2022unsupervised}, and TF-C \cite{zhang2022self}, all carary the risk that the variants of the positive sample might be less similar to the anchor sample compared to the series in the negative set. 
For instance, T-Loss \cite{franceschi2019unsupervised} uses a subseries as a positive sample for a given anchor sample. In~situations where there is a level shift in the anchor sample, the defined positive sample may be less similar to the anchor sample compared to the series in the negative set, where no level shift exists. 
TS-TCC \cite{eldele2021time} uses augmentation techniques such as permutation which carries the same risk. i.e., the permutation of the anchor sample may be very similar to a series in the negative set. 
Figure \ref{fig:tcc-augmentation} shows an example where augmentation techniques proposed for TS-TCC, using jittering and permutation, produce augmented series that are different (dissimilar under Dynamic Time Warping (DTW) distance \cite{sakoe1972dynamic}) from the original series.
The original series of class 0 is more similar to the augmented series of class 2 than to its own augmentation.
Additionally, the augmented series of class 0 and 1 are quite dissimilar to their original series. 
This represents a failure to generate augmentations that are meaningfully similar to the originals while also sufficiently different to allow the creation of useful representations.

For this reason, we propose Series2Vec, a novel self-supervised method inspired by contrastive learning that instead uses learning similarity as its self-supervised task.
Our model utilizes time series similarity measures to assign the target output for the encoder loss, providing a different type of implicit bias that is more suitable for time series analysis than existing pretext methods (pretext refers to the unsupervised task used to generate supervision signals for the target task). This method of creating representations in time series data offers a new and more effective approach to implicit bias encoding. 

This method simply aims to provide similar representations for time series that are similar to each other in the original feature space and dissimilar representations for the time series that are far from each other---
\begin{multline}
    Sim_T(\mathbf{x_i,x_j})<Sim_T(\mathbf{x_i,x_k})\implies \\
    Sim_r(\mathbf{E_T(x_i)},\mathbf{E_T(x_j)})<Sim_r(\mathbf{E_T(x_i)}),\mathbf{E_T(x_k)}).
\end{multline} 
where $Sim_T$ is a relevant similarity measure in the time domain, $Sim_r$ is a relevant similarity measure in the representation domain, $\mathbf{E_T}$ is the function from time series to their representations and $\mathbf{x_i}$, $\mathbf{x_j}$ and $\mathbf{x_k}$ are time series. Since frequency information in time series can be of great importance and is a different/additional source of information, we further extended our model to also learn representations in the frequency domain.

To do so, we propose a novel approach that applies self-attention to each representation within the batch during training. The self-attention mechanism enforces the network to learn similar representations for all similar time series within each batch. One crucial insight motivating this work is the importance of consistency of the targets, not just their correctness, which enables the model to focus on modeling the sequential structure of time series. Our approach draws inspiration from the contrastive learning method for self-supervised representation learning; however, Series2Vec benefits from the similarity prediction loss over time series to represent their structure. Notably, it achieves this without the need for hand-crafted data augmentation.

Additionally, we demonstrate that similarity-based representation learning can be used as a complementary technique with other methods such as self-prediction and contrastive learning to enhance the performance of time series analysis. 

In summary, the main contributions of this work are as follows:
\begin{itemize}
    \item A novel self-supervised learning framework (Series2Vec) is proposed for time series representation learning, inspired by contrastive learning.
    \item A time series similarity measure-based pretext is proposed to assign the target output for the encoder loss, providing a more suitable implicit bias for time series analysis.
    \item A novel approach is introduced that applies order-invariant self-attention to each representation during training, effectively enhancing the preservation of similarity in the representation domain.
    \item The Series2Vec framework was evaluated extensively on nine real-world time series datasets, along with the UCR/UEA archive, and displayed improved results compared to existing SOTA self-supervised methods. It is also evaluated when fused with other representation learning models. 
\end{itemize}

\section{Related Work} \label{A:Related-Works}
Recent advances in self-supervised learning have focused on learning representations through pretext tasks, such as solving jigsaw puzzles \cite{noroozi2016unsupervised}, image colorization \cite{zhang2016colorful}, and predicting image rotation \cite{gidaris2018unsupervised} in the computer vision domain. In the NLP domain, self-supervised models like BERT \cite{devlin2018bert}, and GPT-3 \cite{brown2020language} have also been successful in learning meaningful representations of language. However, these methods rely on heuristics that may limit the generality of the learned representations.
Contrastive learning methods have emerged as an alternative to address this issue, leveraging augmented data to learn invariant representations, such as SimCLR \cite{chen2020simple} in computer vision and ALBERT\cite{lan2019albert} in NLP.

Self-supervised learning for time series classification can mainly be divided into two groups: contrastive learning and self-prediction. This section delves into these approaches. Additionally, a literature review on time series similarity measures has been conducted and is available in Appendix~\ref{App:Related_works} for those interested.

\subsection{Contrastive Learning}
Contrastive learning involves model learning to differentiate between positive and negative time series examples. Scalable Representation Learning (SRL) \cite{franceschi2019unsupervised} and Temporal Neighborhood Coding (TNC) \cite{tonekaboni2021unsupervised} apply a subsequence-based  sampling and assume that distant segments are negative pairs and neighbor segments are positive pairs. TNC takes advantage of the local smoothness of a signal’s generative process to define neighborhoods in time with stationary properties to further improve the sampling quality for the contrastive loss function. 
TS2Vec \cite{yue2022ts2vec} uses contrastive learning to obtain robust contextual representations for each timestamp in a hierarchical manner. It involves randomly sampling two overlapping subseries from input and encouraging consistency of contextual representations on the common segment. The encoder is optimized using both temporal contrastive loss and instance-wise contrastive loss. 

In addition to the subsequence-based methods, there are also other models such as Time-series Temporal and Contextual Contrasting (TS-TCC) \cite{eldele2021time}, Mixing up Contrastive Learning (MCL)~\cite{wickstrom2022mixing}, and Bilinear Temporal-Spectral Fusion (BTSF) \cite{yang2022unsupervised} that employ instance-based sampling. TS-TCC uses weak and strong augmentations to transform the input series into two views and then uses a temporal contrasting module to learn robust temporal representations. The contrasting contextual module is then built upon the contexts from the temporal contrasting module and aims to maximize similarity among contexts of the same sample while minimizing similarity among contexts of different samples \cite{eldele2021time}. BTSF uses simple dropout as the augmentation method and aims to incorporate spectral information into the feature representation \cite{yang2022unsupervised}. Similarly, Time-Frequency Consistency (TF-C) \cite{zhang2022self} is a self-supervised learning method that leverages the frequency domain to achieve better representation. It proposes that the time-based and frequency-based representations, learned from the same time series sample, should be more similar to each other in the time-frequency space compared to representations of different time series samples.

\subsection{Self-Prediction}
 The primary objective of self-prediction-based self-supervised models is to reconstruct the input data. Studies have explored using transformer-based self-supervised learning methods for time series classification, following the success of models like BERT~\cite{devlin2018bert}. BErt-inspired Neural Data Representations (BENDER)\cite{kostas2021bendr} uses the transformer structure to model EEG sequences and shows that it can effectively handle massive amounts of EEG data recorded with differing hardware. Another study, Voice-to-Series with Transformer-based Attention (V2Sa)\cite{yang2021voice2series}, utilizes a large-scale pre-trained speech processing model for time series classification.

Transformer-based Framework (TST)\cite{zerveas2021transformer} adapts vanilla transformers to the multivariate time series domain and uses a self-prediction-based self-supervised pre-training approach with masked data. The pre-trained models are then fine-tuned for downstream tasks such as classification and regression. These studies demonstrate the potential of using transformer-based self-supervised learning methods for time series classification.

\section{Method}
This section begins by formulating the problem of self-supervised time series representation learning. We then introduce our proposed Series2Vec model architecture, which is designed to effectively learn representations from time series data. We also explain the similarity measures that we use in our approach and how they contribute to the effectiveness of our method. Finally, we describe our pretext method for self-supervised time series representation learning i.e., self-supervised similarity-preserving. We outline our approach for defining a model that can effectively capture and preserve the underlying similarity within the data. 
\subsection{Problem Definition} 
In this study, our aim is to tackle the problem of learning a nonlinear embedding function that can effectively map each time series $\mathbf{x_i}$ from a given dataset $X$ into a condensed and meaningful representation $r_i \in \mathbb{R}^K$, where $K$ denotes the desired representation dimension. The dataset $X$ comprises $n$ samples, specifically $X=\left\{\mathbf{x_1},\mathbf{x_2},...,\mathbf{x_n}\right\}$, where each $\mathbf{x_i}$ represents a $d_x$-dimensional time series of length $L$. We denote that $\mathbf{x_i} \equiv \mathbf{x_i^T}$ represents an input time series sample, and $\mathbf{x_i^F}$ represents the discrete frequency spectrum of $\mathbf{x_i}$. We define $r_i^T$ as the representation of $\mathbf{x_i}$ sample in the time domain, and $r_i^F$ as the representation of $\mathbf{x_i}$ in the frequency domain, and $r_i$ is the concatenation of $[r_i^T,r_i^F]$. To evaluate the quality of our learned representation $\mathbf{r}=\{r_1,r_2..,r_n\}$, we consider two scenarios based on the availability of labeled data: \textit{Linear Probing} and \textit{Fine-Tuning}.

\subsubsection*{Linear Probing} We assume access to a large volume of unlabeled data $X^U = \left\{\mathbf{x_i} \lvert i=1,...,n\right\}$, along with a smaller subset of labeled data $X^L = \left\{(\mathbf{x_i}, y_i) \lvert i=1,...,m\right\}$ samples ($m \ll n$). Each sample in $X^L$ is associated with a label $y_i \in {1, . . . , C}$, where $C$ represents the number of classes.
First, we pre-train a model without using labels through a self-supervised pretext task. Once the pre-training is complete, we freeze the encoder and add a linear classifier on top of the pre-trained model's output or intermediate representations. This linear classifier can be implemented as a linear layer or logistic regression. The linear classifier is subsequently trained on a downstream task, typically a classification task, utilizing the pre-trained representations as inputs. Linear probing serves as an evaluation method to assess the quality of the learned representations.

\subsubsection*{Fine-Tuning} We assume that the dataset $X$ is fully labeled, denoted as $X = \left\{(\mathbf{x_i}, y_i) \lvert i=1,...,n\right\}$. Each sample in $X^L$ is associated with a label $y_i \in \left\{1, . . . , C\right\}$, where $C$ represents the number of classes.
We investigate whether leveraging similarity-based representation learning for initialization provides advantages compared to randomly initializing a supervised model. To examine this, we first pre-train the model without using labels through a self-supervised pretext task. Afterward, we train (fine-tune) the entire model for a few epochs using the labeled dataset in a fully supervised manner.
\subsection{Model Architecture} 

\begin{figure*}
    \centering
    \includegraphics[trim=0.5cm 5.5cm 4cm 1cm, width=1\textwidth]{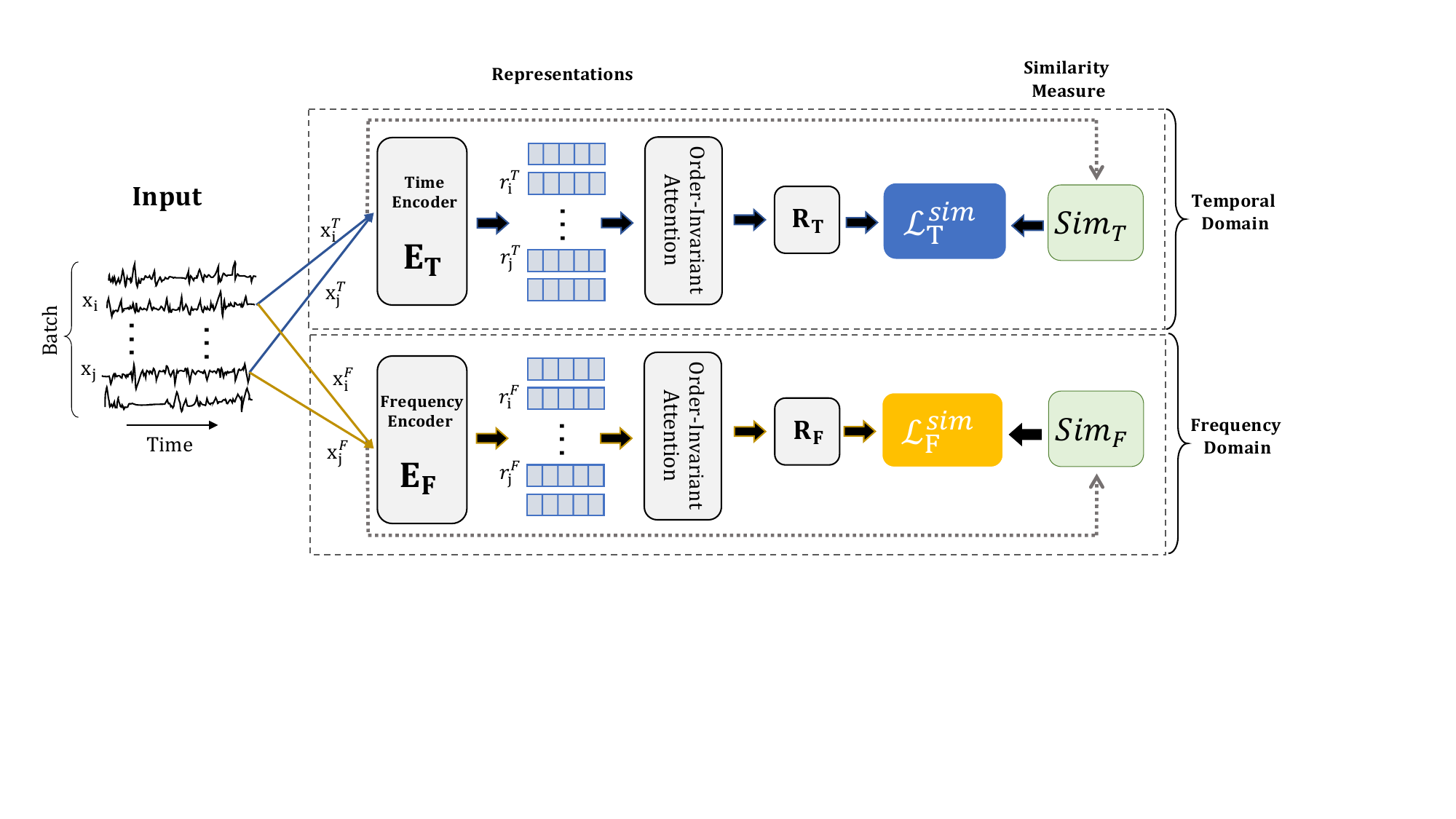}
    \caption{Architecture of Series2Vec. The top module learns the representations in the temporal domain and the bottom module learns the representations in the frequency domain.}
    \label{fig:Series2Vec}
    \vspace{-0.5cm}
\end{figure*} 
The overall architecture of Series2Vec is shown in Figure~\ref{fig:Series2Vec}. The Series2Vec model architecture proposed in this work is designed to handle both univariate and multivariate time series inputs. However, for the purpose of simplicity, we will focus on illustrating the model using univariate time series in the following descriptions. As shown in Figure~\ref{fig:Series2Vec} the model comprises four main components: a time encoder ($\mathbf{E_t}$), a frequency encoder ($\mathbf{E_f}$), a similarity measuring functions for time and frequency (section \ref{sec:Soft-DTW}), and an similarity-preserving loss function (section \ref{Sec:Similarity}). The encoder blocks map the input time series data into condensed and meaningful representations in both time and frequency domains. A similarity measuring function calculates the similarity between pairs of input series, providing a quantitative measure of their resemblance. To optimize the encoder blocks, a similarity-preserving loss function is employed. This loss function guides the learning process, encouraging the encoder blocks to learn representations that preserve the similarity relationships between different samples in the dataset in both time and frequency domains.

For a given input time series sample, denoted as $\mathbf{x_i}$, we obtain its corresponding frequency spectrum, $\mathbf{x^F_i}$, through a transform operator such as the Fourier Transformation \cite{FFT1988}. The frequency spectrum captures universal frequency information within the time series data, which has been widely acknowledged as a key component in classical signal processing \cite{FFT1988}. Furthermore, recent studies have demonstrated the potential of utilizing frequency information to enhance self-supervised representation learning for time series data \cite{zhang2022self,yang2022unsupervised}.

The time-domain input $\mathbf{x^T_i}$ and the frequency-domain input $\mathbf{x^F_i}$ are separately passed into the time and frequency encoders to extract features. The feature extraction process is as follows:
\begin{equation} 
 r^T_i=\mathbf{E_T}(\mathbf{x^T_i},\theta_T), \quad  r^F_i=\mathbf{E_F}(\mathbf{x^F_i},\theta_F)
\end{equation}
where $\theta_T$ and $\theta_F$ represent the parameters of the time and frequency encoders, respectively. The encoded representations of $\mathbf{x_i}$ are denoted as $r^T_i\in \mathbb{R}^K$ and $r^F_i\in \mathbb{R}^K$. Following the established setup outlined in previous works (e.g., \cite{foumani2021disjoint,Foumani2023}), we adopt Disjoint convolutions for encoding both temporal and spectral features. These convolutions efficiently capture the temporal and spatial features \cite{foumani2021disjoint}.
To ensure consistent representation sizes, we employ max pooling at the end of the encoding network. This choice guarantees the scalability of our model to different input lengths.

\subsection{Similarity Measuring Function} 
\label{sec:Soft-DTW}

Soft-DTW \cite{cuturi2017soft} is employed as the similarity function in time domain.
It was proposed as an alternative to DTW and we used it due to the availability of an efficient GPU implementation of Soft-DTW\footnote{\url{https://github.com/Maghoumi/pytorch-softdtw-cuda}}, allows our proposed method to be more efficient, scale, and run faster on large time series datasets.
The distance calculated by Soft-DTW is a continuous and differentiable function. 
The formulation for Soft-DTW distance is given by
\vspace{-0.4cm}
\begin{equation}
\mathcal{S}_T(\mathbf{x^T_i},\mathbf{x^T_j}) = \min_{\pi} \sum_{i=1}^{L}||\mathbf{x^T_i}-\mathbf{x^T_{j,\pi(i)}}||^2 e^{-\frac{\alpha}{2}||i-\pi(i)||^2}
\label{eq:soft-dtw}
\vspace{-0.4cm}
\end{equation}

Where $\mathbf{x^T_i}$ and $\mathbf{x^T_j}$ are the two time series being compared, $L$ is the length of the time series, and $\pi$ is a warping path. The warping path is defined as a function that maps each index of one time series to a corresponding index in the other time series. The goal is to find the warping path that minimizes the sum of the squared distances between the corresponding elements of the two time series. The parameter $\alpha \in [0,1]$ controls the degree of alignment between the two time series. 
Smaller values of $\alpha$ result in a more accurate alignment, while larger values lead to a more robust alignment.
It is worth noting that setting $\alpha=0$ makes Soft-DTW and DTW equivalent. 

For the similarity function in the frequency domain, we use the Euclidean distance as unlike the temporal domain where Soft-DTW is employed, the concept of time warping does not apply directly to the frequency domain. The Euclidean distance between two input series $\mathbf{x^F_i}$ and $\mathbf{x^F_j}$ can be calculated as follows:
\begin{equation}
\mathcal{S}_F(\mathbf{x^F_i},\mathbf{x^F_j}) = \sqrt{\sum_{k=1}^{M}||\mathbf{x^F_{i,k}}-\mathbf{x^F_{j,k}}||^2}
\label{eq:euc}
\end{equation}
Here, $\mathbf{x^F_i}$ and $\mathbf{x^F_j}$ represent the frequency domain representations of two time series being compared, and $M$ is the number of frequency bins. The Euclidean distance is computed by taking the square root of the sum of squared differences between corresponding frequency components of the two representations.
\subsection{Self-Supervised Similarity-Preserving} 
\label{Sec:Similarity}


Contrastive learning has been successfully used in computer vision and natural language processing due to the strong constraints present in image and text data. In NLP, syntax, and semantics constrain the ordering and meaning of tokens, making it easier to define meaningful variants of the positive samples, e.g., replacing a word with its synonym \cite{brown2020language}. Similarly, images can be analyzed based on the subject matter, and transformations such as scaling, blurring, and rotation can still be used to identify the same subject. However, the wide variety of possible sources and processes in time series data makes it more challenging to apply the same constraints and techniques used in computer vision and NLP for learning representations in time series, all carry the risk that the variants of the positive sample (such as permutation \cite{eldele2021time} and subseries selection \cite{franceschi2019unsupervised}) might be less similar to the anchor sample compared to the series in the negative set.


We propose a novel pretext task that is specifically designed to address the unique challenges and characteristics of time series data. Our task aims to model a different type of implicit bias that is more suitable for time series analysis. Our proposed Series2Vec utilizes a similarity measure to align the target output for the encoder loss. The key question now is what is the best approach that effectively captures and preserves this similarity.

To simplify the explanation, we will focus on the time domain and omit the frequency domain. Let's assume that $r_i$ and $r_j$ are the representation vectors for input time series $\mathbf{x_i}$ and $\mathbf{x_j}$, respectively. Our main objective is to learn similar representations for all similar time series within each batch. To accomplish this, we leverage transformers and make use of the order-invariant property of self-attention mechanisms.

In our approach, each time series within each batch functions as a query and attends to the keys of the other samples in the batch in order to construct its representation. This process allows the representation we seek to capture and aggregate all the relevant information from the input representations of the entire batch. By employing the transformer's architecture and utilizing self-attention, we aim to generate comprehensive representations that encapsulate the pertinent characteristics and similarities among the input time series samples. 

To the best of our knowledge, our work is the first to introduce the concept of feeding each time series as an input token to transformers in order to learn similarity-based representations. In our approach, we utilize transformers to model the relationships and interactions between the time series within the batch. By treating each time series as a separate input token, we enable the model to capture the fine-grained similarities between different series. 

Specifically, transformers map a query and a set of key-value pairs to an output. For an input batch representation, $\mathbf{R} = \left\{r_1,r_2,...,r_B\right\}$ where $B$ is the batch size, self-attention computes an output series $\mathbf{Z} =\left\{z_1,z_2,...,z_B\right\}$ where $z_i\in \mathbb{R}^{d_z} $ and is computed as a weighted sum of input elements:
\vspace{-0.4cm}
\begin{equation} 
\label{e1}
z_i=\sum_{j=1}^B \alpha_{i,j}(r_j W^V)
\vspace{-0.3cm}
\end{equation}
Each coefficient weight $\alpha_{i,j}$ is calculated using a softmax function:
\vspace{-0.3cm}
\begin{equation}
\label{e2}
\alpha_{i,j}=\frac{exp(e_{ij})}{\sum_{k=1}^B exp(e_{ik})}
\end{equation}
where $e_{ij}$ is an attention weight from representations  $j$ to $i$ and is computed using a scaled dot-product:
\vspace{-0.3cm}
\begin{equation}
\label{e3}
e_{ij}=\frac{(r_i W^Q)(r_j W^K)^T}{\sqrt{d_z}}
\end{equation}
The projections $W^Q, W^K, W^V \in \mathbb{R}^{K \times d_z}$ are parameter matrices and are unique per layer. Instead of computing self-attention once, Multi-Head Attention (MHA) \cite{vaswani2017attention} does so multiple times in parallel, i.e., employing $h$ attention heads. 


Assuming $z_i, z_j \in \mathbb{R}^{d_z}$ are the output vectors of transformers for input representation $r_i$ and $r_j \in \mathbb{R}^{K}$, respectively. The pretext objective we have defined aims to minimize the following loss function:
\begin{equation}
\label{eq:loss}
    \mathcal{L}_T^{\text{sim}} = \text{smooth}_{L_1} (R_T(z_i, z_j), Sim_{T} \mathbf{(x_i, x_j)})
\end{equation}

The equation \ref{eq:loss} represents the similarity loss between the encoded representations \(z_i\) and \(z_j\) using our encoder. It is calculated the smooth $L_1$ loss~\cite{girshick2015fast} between the similarity \(R_T(z_i, z_j)\) and similarity function ${Sim_T}\mathbf{(x_i, x_j)}$. The smooth $L_1$ loss is defined as:
\vspace{-0.3cm}
\begin{equation}
\text{smooth}_{L_1}(x) = 
\begin{cases} 
0.5x^2 & \text{if } |x| < 1 \\
\lvert x \rvert - 0.5 & \text{otherwise}
\end{cases}
\vspace{-0.3cm}
\end{equation}

We chose smooth L1 loss because the literature shows it is less sensitive to outliers compared to MSE loss, and in certain scenarios, it prevents the issue of exploding gradients \cite{girshick2015fast}. We also found experimentally that it performs better than MSE loss. The similarity \(R_T(z_i, z_j)\) is computed by taking the dot product of the encoded vectors \(z_i\) and \(z_j\). The similarity ${Sim_T}\mathbf{(x_i, x_j)}$ is calculated between the time series $\mathbf{x_i}$ and $\mathbf{x_j}$ using equation \ref{eq:soft-dtw}.

In our model, we follow the same process for the frequency domain. The loss function is defined as follows:
\vspace{-0.3cm}
\begin{equation}
\label{eq:loss_f}
\mathcal{L}_F^{\text{sim}} = \text{smooth}_{L_1} (R_F(z^F_i, z^F_j), \text{Sim}_F(\mathbf{x^F_i, x^F_j)})
\end{equation}
Here, the similarity $\text{Sim}_F(\mathbf{x^F_i, x^F_j})$ is calculated between $\mathbf{x^F_i}$ and $\mathbf{x^F_j}$ using equation~\ref{eq:euc}. The total loss is then calculated as:
\vspace{-0.2cm}
\begin{equation}
\mathcal{L}_{\text{Total}} = \mathcal{L}_T^{\text{sim}} + \mathcal{L}_F^{\text{sim}}
\vspace{-0.2cm}
\end{equation}

Training the encoder using $\mathcal{L}_{\text{Total}}$ loss function that is based on a time series-specific similarity measure enabled the model to learn a representation of the input data that effectively captures the similarities between the series in each batch. Additionally, time series-specific similarity measures are able to align and compare time series with different time steps and lengths by warping the time axis, making the loss function robust to non-linear variations in the data. This makes the model more robust and less sensitive to small variations in the data, which in turn improves its ability to generalize to unseen time series data. Furthermore, by training the model with a loss function that is based on time series-specific similarity measures, the model is exposed to a wide range of time series variations, such as different time steps, lengths, and irregular intervals, which allows it to learn the underlying patterns in the data that are specific to time series. Time series-specific similarity measures like Dynamic Time Warping (DTW) can handle irregular time intervals, non-stationary time series, and variable-length time series, which can be beneficial when training the model with time series that have these characteristics.

The primary focus of our proposed pretext model is to leverage the similarity information between time series, without being limited by the quality of a specific similarity measure. This allows for flexibility in the choice of similarity measure, as any time series similarity measure can be plugged into the model and used to learn representations. In this paper, we chose a time series-specific similarity measure, Soft-DTW \cite{cuturi2017soft} (please refer to section \ref{sec:Soft-DTW} for the reason why we used this similarity measure). Clearly, our proposed model is not limited to specific similarity measures and can be easily extended to incorporate other similarity measures as well.

\section{Experimental Results}
This section presents the experimental results of our study, focusing on the performance evaluation of the Series2Vec model in a downstream task of time series classification. The experiments are divided into three main parts: 1) linear probing, 2) fine-tuning, and 3) ablation study. Our primary objective is to assess the effectiveness of the learned representation in accurately classifying time series data and to compare Series2Vec performance against other state-of-the-art models. Additional experiments on the UCR/UEA archive are provided in the Appendix~\ref{A:UCR/UEA} due to space constraints. Here we evaluate models on commonly used datasets in the representation learning literature. 

\subsection{Datasets}

To evaluate the performance of our model, we utilize a total of nine publicly available datasets that have been previously used in the literature for time series representation learning \cite{foumani2023deep}. These datasets cover various domains, such as epileptic seizure prediction \cite{Epilepsy}, sleep stage classification \cite{Sleep}, and human activity recognition datasets such as \cite{HAR}, PAMAP2 \cite{PAMAP2-2012}, Skoda \cite{skoda2012}, USC-HAD \cite{USC-HAD2012}, Opportunity \cite{Opportunity}, WISDM \cite{WISDM}, and WISDM2 \cite{WISDM2}. The details of each dataset are presented in Appendix~\ref{Apx:datasets}.   
\subsection{Evaluation Procedure and Parameter Setting}
We evaluate model performance using classification accuracy as the main metric following the literature in time series classification. Models are ranked based on their accuracy per dataset, with the highest accuracy receiving a rank of 1 and the lowest rank assigned to the worst performer. In the case of ties, the average rank is calculated. The final step is to compute the average rank of each model across all datasets. 
This gives a direct general assessment of all the models: the lowest rank corresponds to the method that is the most accurate on average. For the statistical test, we used the Wilcoxon signed-rank test \cite{demvsar2006statistical}.

In our experiments, the Series2Vec model employed two layers of temporal and spatial convolutions \cite{foumani2021disjoint} to encode temporal and spectral features. The model utilized 16 filters per layer in the temporal and spatial convolution layers. During training, a batch size of 64 was used, and the Adam optimization algorithm \cite{kingma2014adam} was employed. To prevent overfitting, an early stopping method based on the validation loss was implemented. The model is pre-trained for 100 epochs, and logistic regression is then applied to the representations for linear probing.

Similar to the transformer-based model for multivariate time series classification (TST) \cite{zerveas2021transformer} and the default transformer block \cite{vaswani2017attention}, in our experiments we utilized eight attention heads to capture the diverse features from the input time series. The transformer encoding dimension was set to $d_m=64$, and the feed-forward network (FFN) in the transformer block expanded the input size by 4x before projecting it back to its original size.

The Soft-DTW's parameter $\alpha$, which determines the level of alignment between the two time series, is set to $0.1$ as per the original paper's recommendation \cite{cuturi2017soft}.
\subsection{Linear Probing}
\subsubsection{Comparison with Baseline Approaches}
\label{sec: baseline}
In order to evaluate the effectiveness of our approach, we conducted extensive comparison against six state-of-the-art self-supervised methods for time series, including TS2Vec \cite{yue2022ts2vec}, TS-TCC \cite{eldele2021time}, TNC  \cite{tonekaboni2021unsupervised}, TF-C \cite{zhang2022self}, MCL \cite{wickstrom2022mixing} and TST \cite{zerveas2021transformer}. To ensure a fair comparison, we used publicly available code for the baseline methods.


\begin{table}[]
\centering
\footnotesize
\setlength{\tabcolsep}{1pt}
\setlength\extrarowheight{3pt}
\caption{Comparing self-supervised models: An analysis of average accuracy scores for Series2Vec, TS2Vec, TS-TCC, TNC, MCL, TF-C and TST.}
\begin{tabular}{lccccccc} \hline
\textbf{Datasets} & \textbf{Series2Vec} & \textbf{TS2Vec} & \textbf{TS-TCC} & \textbf{TNC} & \textbf{MCL} & \textbf{TF-C} & \textbf{TST} \\ \hline \hline

WISDM2 & $\textbf{64.70} \pm \text{\tiny{0.34}}$ & $62.50 \pm \text{\tiny{0.49}}$ & $64.36 \pm \text{\tiny{1.02}}$ & $63.75 \pm \text{\tiny{0.9}}$ & $63.84 \pm \text{\tiny{0.98}}$ & $63.35 \pm \text{\tiny{2.05}}$ & $61.75 \pm \text{\tiny{1.34}}$ \\

PAMAP2 & $\textbf{81.99} \pm \text{\tiny{0.41}}$ & $77.33 \pm \text{\tiny{0.52}}$ & $62.33 \pm \text{\tiny{1.23}}$ & $67.25 \pm \text{\tiny{1.09}}$ &$61.93 \pm \text{\tiny{1.74}}$ & $40.32 \pm \text{\tiny{1.74}}$ & $52.80 \pm \text{\tiny{1.15}}$ \\

USC-HAD & $\textbf{56.12} \pm \text{\tiny{1.34}}$ & $47.28 \pm \text{\tiny{4.17}}$ & $53.89 \pm \text{\tiny{3.43}}$ & $51.19 \pm \text{\tiny{2.05}}$ & $50.45 \pm \text{\tiny{3.60}}$ & $48.51 \pm \text{\tiny{2.83}}$ & $46.57 \pm \text{\tiny{5.29}}$ \\

Sleep & $\textbf{80.62} \pm \text{\tiny{3.59}}$ & $77.38 \pm \text{\tiny{4.74}}$ & $80.59 \pm \text{\tiny{1.82}}$ & $75.71 \pm \text{\tiny{4.93}}$ & $74.82 \pm \text{\tiny{4.16}}$ & $73.82 \pm \text{\tiny{6.63}}$ & $74.93 \pm \text{\tiny{5.68}}$  \\

Skoda & $\textbf{98.65} \pm \text{\tiny{0.11}}$ & $98.09 \pm \text{\tiny{0.44}}$ & $96.39 \pm \text{\tiny{0.52}}$ & $95.39 \pm \text{\tiny{0.27}}$ & $94.04 \pm \text{\tiny{0.53}}$ & $96.40 \pm \text{\tiny{0.67}}$ & $86.95 \pm \text{\tiny{1.41}}$ \\

Opp & $\textbf{86.56} \pm \text{\tiny{0.42}}$ & $85.00 \pm \text{\tiny{1.56}}$ & $82.90 \pm \text{\tiny{1.68}}$ & $83.81 \pm \text{\tiny{4.73}}$ & $82.54 \pm \text{\tiny{0.85}}$ & $83.06 \pm \text{\tiny{1.25}}$ & $82.81 \pm \text{\tiny{4.61}}$ \\

WISDM & $\textbf{81.17} \pm \text{\tiny{0.27}}$ & $80.23 \pm \text{\tiny{2.31}}$ & $76.16 \pm \text{\tiny{2.39}}$ & $73.16 \pm \text{\tiny{5.47}}$ & $74.06 \pm \text{\tiny{2.12}}$ & $72.08 \pm \text{\tiny{3.00}}$ & $62.88 \pm \text{\tiny{6.54}}$ \\ 

Epilepsy & $\textbf{97.61} \pm \text{\tiny{0.03}}$ & $97.52 \pm \text{\tiny{0.18}}$ & $96.83 \pm \text{\tiny{0.12}}$ & $95.83 \pm \text{\tiny{0.15}}$ & $96.91 \pm \text{\tiny{0.13}}$ & $96.73 \pm \text{\tiny{0.33}}$ & $82.61 \pm \text{\tiny{0.24}}$ \\

UCI-HAR & $\textbf{94.77} \pm \text{\tiny{0.13}}$ & $93.76 \pm \text{\tiny{1.17}}$ & $89.21 \pm \text{\tiny{0.24}}$ & $90.21 \pm \text{\tiny{0.27}}$ & $88.46 \pm \text{\tiny{2.84}}$ & $81.09 \pm \text{\tiny{1.46}}$ & $86.21 \pm \text{\tiny{2.31}}$ \\ \hline

\textbf{Average} & \textbf{82.47}  & 79.90 & 78.07 & 77.37 & 76.33 & 72.81&  70.83 \\
\textbf{Rank} & \textbf{1} & 3 & 3.33 & 4 & 4.78 & 5.44 & 6.44 \\ \hline
\end{tabular}
\label{tab:Comparing}
\end{table}

Table \ref{tab:Comparing} presents the average accuracy of Series2Vec over five runs, along with other state-of-the-art self-supervised models, for the purpose of comparison. 
The number in bold for each dataset represents the highest accuracy achieved for that dataset.
The last row in Table \ref{tab:Comparing} shows the rank of each model across all nine datasets. The results presented in Table \ref{tab:Comparing} indicate that our model, Series2Vec, achieves the highest average rank of 1 (which is significantly more accurate than other models) and the highest average accuracy of 82.47 among all self-supervised models. The second most accurate model, TS2Vec, obtains an average rank of 3 and an average accuracy of 79.90. TS-TCC follows closely with an average accuracy of 78.07. TST is the worst-performing model with an average accuracy of 70.83.

\subsubsection{Low-Label Regimes}

\begin{figure*}
    \centering
    \includegraphics[trim=5cm 2.5cm 4.5cm 4cm, width=1\textwidth]{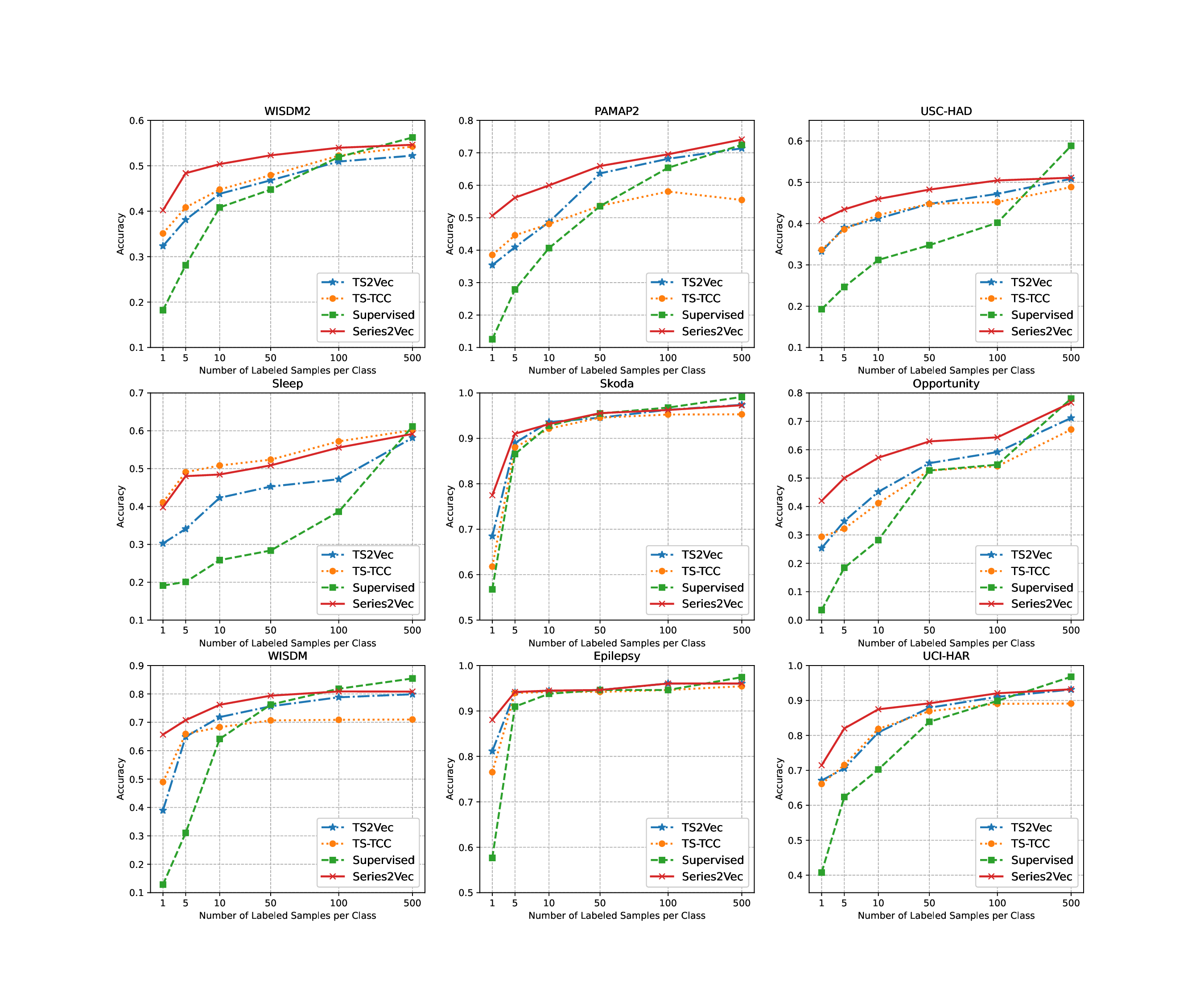}
    \caption{Comparison of Linear Probing with Series2Vec, TS2Vec, TS-TCC and Supervised on all nine datasets. The x-axis represents the number of labeled samples per class, while the y-axis represents the corresponding accuracy achieved by each approach}
    \label{fig:Low-Label}
    \vspace{-0.2cm}
\end{figure*}

We conducted a comparison between three self-supervised models (Series2Vec, TS2Vec, and TS-TCC) and a supervised model in a low-labeled data regime. The TNC, TF-C, MCL, and TST models were excluded from the comparison due to their significantly lower accuracy compared to the other models. Figure \ref{fig:Low-Label} demonstrates that our proposed Series2Vec model consistently outperforms both the supervised model and other representation learning models (except for one dataset -Sleep- in comparison to TS-TCC) when the number of labeled data points is limited to less than 50. Note each subfigure here shows the results for one dataset. This indicates the promising performance of Series2Vec models in scenarios where data scarcity is a challenge. Notably, the Series2Vec models exhibit consistent performance across all datasets, which adds to the reliability of our findings. It is important to highlight that TS-TCC, which uses augmentation techniques, performs similarly to our model on Sleep datasets. Sleep dataset consists of EEG signals, and enhancing the model's ability to handle noise would be especially beneficial in this scenario.

\subsection{Pre-Training}
Our objective here is to evaluate the effectiveness of our model in the pre-training phase. Table \ref{tab:Initialization} presents the classification accuracy results for different datasets, comparing the performance of a model with random initialization and pre-trained Series2Vec. The table shows that using pre-trained Series2Vec leads to an average improvement of 1\% in accuracy compared to the random initialization. Significant improvements are observed in specific datasets, such as WISDM2, PAMAP2, and WISDM. For WISDM2, Series2Vec achieves an accuracy gain of 2.35\% compared to the random initialization. Similarly, for PAMAP2 and WISDM, the accuracy gains are 3.03\% and 1.51\% respectively, validating the effectiveness of utilizing similarity-based methods for enhanced learning and improved time series classification.

\begin{table}[]
\centering
\footnotesize
\setlength{\tabcolsep}{1.5pt}
\setlength\extrarowheight{4pt}
\vspace{-0.5cm}
\caption{Comparison of Classification Accuracy between Random Initialization and Pre-Trained Series2Vec}

\begin{tabular}{l|ccccccccc|r} \hline
Datasets & WIS2 & PAM2 & USC & Sleep & Skoda & Opp & WIS & Ep & HAR & \textbf{Average} \\ \hline \hline
\textbf{Random} & 65.91 & 75.10 & 56.26 & 82.20 & 99.16 & 89.34 & 84.69 & 98.17 & 96.06 & \textbf{82.99} \\
\textbf{Series2Vec} & \textbf{68.26} & \textbf{78.13} & \textbf{56.89} & \textbf{82.78} & \textbf{99.37} & \textbf{89.60} & \textbf{86.20} & \textbf{98.35} & \textbf{96.40} & \textbf{84.00} \\\hline 
\end{tabular}
\label{tab:Initialization}
\vspace{-0.25cm}
\end{table}

\subsection{Ablation Study}
\textbf{Component Analysis: } To assess the effectiveness of the proposed components in Series2Vec, we conducted a comparison between the Series2Vec model and three variations, as presented in Table \ref{tab:Component}. The variations are as follows: (1) \textbf{w/o Attention}, where the transformer block is removed; (2) \textbf{w/o Spectral}, where only the temporal domain is used as input feature; and (3) \textbf{w/o Temporal}, where the frequency of the input series is solely utilized to generate the representation.

\begin{table}[]
\centering
\footnotesize
\setlength{\tabcolsep}{2 pt}
\setlength\extrarowheight{2 pt}
\vspace{-0.4cm}
\caption{Series2Vec Ablation Study: Component Analysis}
\begin{tabular}{l|ccccccccc|c} \hline
\textbf{Component} & WIS2 & PAM2 & USC & Sleep & Skoda & Opp & WIS & Ep & HAR & Average \\ \hline \hline
\textbf{w/o Attention} & 59.83 & 70.64 & 36.84 & 61.21 & 93.25 & 82.97 & 74.70 & 91.00 & 90.03 & 73.38 ($\downarrow$ 9.08) \\
\textbf{w/o Spectral} & 60.83 & 78.64 & 38.84 & 70.21 & 96.25 & 85.97 & 79.70 & 97.00 & 92.60 & 77.78 ($\downarrow$ 4.68) \\
\textbf{w/o Temporal} & 63.16 & 71.88 & 45.82 & 63.36 & 98.42 & 86.45 & 80.54 & 96.65 & 93.62 & 77.76 ($\downarrow$ 4.70) \\  \hline
\textbf{Series2Vec} & \textbf{64.7} & \textbf{81.99} & \textbf{56.12} & \textbf{80.62} & \textbf{98.65} & \textbf{86.56} & \textbf{81.17} & \textbf{97.61} & \textbf{94.77} & \textbf{82.47} \\ \hline
\end{tabular}
\label{tab:Component}
\vspace{-0.25cm}
\end{table}

As shown in Table \ref{tab:Component}, the inclusion of order-invariant self-attention has a significant impact on the model's accuracy, thereby validating our approach, which employs it to ensure that the model attends to similar series in the batch for a given time series. Furthermore, we observed that in datasets recorded with a low sampling rate such as WISDM2, Skoda, WISDM, and UCI-HAR, employing the frequency domain improves the model's performance. Low sampling may make it difficult for the model to capture fine-grained temporal patterns in the data. However, frequency-based representations derived from the FFT can capture information about the underlying periodicity and spectral content of the signal. 

\textbf{Complementary Loss Function}
We evaluate our similarity preserving loss ($\mathcal{L}_{Sim}$) performance in combination with other methods such as self-prediction loss ($\mathcal{L}_{SP}$) used in TST and contrastive loss ($\mathcal{L}_{Cons}$) employed in TS-TCC. Table~\ref{tab:Complementary} showcases the average accuracy of five runs for different combinations of similarity, contrastive, and self-prediction loss on all nine datasets. 
Notably, we find that the similarity loss surpasses the individual performance of self-prediction loss in TST and contrastive loss in TS-TCC. Additionally, the combination of self-prediction and similarity-preserving learning yields superior results compared to the combination of contrastive and similarity loss. This suggests that self-prediction and similarity learning capture distinct implicit biases, and their fusion leads to enhanced performance in time series analysis.

\begin{table}[]
\centering
\setlength{\tabcolsep}{25 pt}
\setlength\extrarowheight{2pt}
\caption{$\mathcal{L}_{Sim}$ as Complementary Loss Function}
\begin{tabular}{lc}
\textbf{Loss Function} & \textbf{Average Accuracy} \\ \hline \hline
$\mathcal{L}_{Sim}$ (Series2Vec) & \textbf{82.47} \\ \hline
$\mathcal{L}_{Cons}$ (TS-TCC) & 78.07 \\
$\mathcal{L}_{SP}$ (TST) & 70.83 \\
$\mathcal{L}_{Sim} + \mathcal{L}_{Cons}$ & 82.61 ($\uparrow$ 0.14) \\
$\mathcal{L}_{Sim} + \mathcal{L}_{SP}$ & 83.45 ($\uparrow$ 0.98) \\
$\mathcal{L}_{Sim} + \mathcal{L}_{SP} +\mathcal{L}_{Cons}$ & \textbf{83.55 ($\uparrow$ 1.08)} \\ \hline
\end{tabular}
\vspace{-0.4cm}
\label{tab:Complementary}
\end{table}

\section{Conclusion}
 This paper proposes a novel self-supervised learning method, Series2Vec, for time series analysis. Series2Vec is inspired by contrastive learning, but instead of using synthetic transformations, it utilizes time series similarity metrics to assign the target output for the encoder loss. This method offers a novel and more effective approach to implicit bias encoding, making it more suitable for time series analysis. The results of the experiments show that Series2Vec outperforms existing methods for time series representation learning. Additionally, our experimental results indicate that Series2Vec performs well in datasets with a limited number of labeled samples. Finally, fusion of Series2Vec with other representation learning models leads to enhanced performance in time series classification. 



\bibliography{bibliography}

\begin{appendices}

\clearpage 
\section{Related Work on Similarity Measures} \label{App:Related_works}
A similarity measure calculates the distance between two time series and the smaller the distance, the more similar the two time series. There have been many similarity measures developed for time series data. 
Time series similarity measures play an important role in almost all time series data mining tasks such as classification, regression, anomaly detection, motif discovery and clustering \cite{tan2021time, petitjean2011global}.
One of the popular similarity measures for comparing a pair of time series is Dynamic Time Warping (DTW)  \cite{sakoe1972dynamic}. 
DTW calculates the distance by aligning the two time series in a non-linear way, allowing more flexible comparison than traditional methods such as Euclidean distance.
It is also robust to shifts and dilation across the time dimension \cite{sakoe1972dynamic}.
This makes DTW useful for comparing time series that may have been recorded at different times or at different frequencies and a valuable tool for many applications \cite{petitjean2011global}.

Its popularity has led to various extensions of the measure.
The weighted DTW (WDTW) \cite{jeong2011weighted} and the recent Amerced DTW (ADTW) \cite{herrmann2023amercing} are variants of DTW that penalize off-diagonal warping paths. 
ADTW was shown to significantly outperform both DTW and WDTW when benchmarked on the univariate UCR time series archive \cite{dau2019ucr}.
The use of DTW as a feature was explored in \cite{kate2016using} where each time series is represented as a vector of DTW distances to each of the examples in the training dataset.
The authors demonstrated the effectiveness of DTW features using a support vector machine (SVM) and concatenated with other features, where their method was more accurate than nearest neighbor classification. 

Since DTW is not differentiable, it cannot be used as a loss function for the neural networks. 
Hence, Soft-DTW \cite{cuturi2017soft} (see Section \ref{sec:Soft-DTW}) was developed to allow DTW to be used as a loss function to train neural networks.
The authors showed that using Soft-DTW as the measure for clustering and forecasting is superior to using DTW \cite{cuturi2017soft}.
Given the benefits of Soft-DTW and the availability of an efficient GPU implementation, Soft-DTW is used as a proof of concept for our work. We will consider the exploration of other similarity measures for time series self-supervised learning as our future work. 

\section{Datasets} \label{Apx:datasets}
We chose these datasets as they are commonly employed in self-supervised representation learning for time series research \cite{eldele2021time, zhang2022self}. The details of each dataset are provided in Table \ref{Tab:Data}. For all datasets except Skoda, we performed subject-wise data splitting, ensuring that the test set comprises at least 20 percent of the data. However, since the Skoda datasets were recorded using only one subject, subject-wise data splitting was not applicable in this case.
\begin{table}[h]
\centering
\setlength{\tabcolsep}{5pt}
\setlength\extrarowheight{2pt}
\caption{Description of datasets used in our experiments.}
\begin{tabular}{l|cccccc} 
\hline
Dataset & \multicolumn{1}{l}{\#Train} & \multicolumn{1}{l}{\# Test} & \multicolumn{1}{l}{\# Subject} & \multicolumn{1}{l}{Length} & \multicolumn{1}{l}{\# Channel} & \multicolumn{1}{l}{\# Class} \\ \hline \hline
WISDM2    & 134,614 & 14,421 & 29 & 40  & 3  & 6 \\
PAMAP2    & 51,192  & 11,590 & 9   & 100 & 52 & 18 \\
USC-HAD   & 46,899 & 9,327 & 14   & 100 & 6 & 12 \\
Sleep     & 25,612 & 8,910 & 20 & 3000 & 1 & 5 \\
Skoda     & 22,587 & 5,646 & 1    & 50 & 64 & 11 \\
Opportunity &15,011 & 2,374 & 4 & 100 & 113 & 18 \\
WISDM     & 11,960 & 5,207      & 13 & 40 & 3 & 6 \\
Epilepsy  & 9,200 & 2,300   & 500 & 178 & 1 & 2 \\
UCI-HAR   & 7,352 & 2,947 & 30 & 128 & 9 & 6 \\
\hline
\end{tabular}
\label{Tab:Data}
\vspace{-0.25cm}
\end{table}

\section{Additional experiments on UCR/UEA} \label{A:UCR/UEA}
\begin{figure}[h]
    \centering
    \begin{subfigure}{\columnwidth}
        \includegraphics[width=\columnwidth]{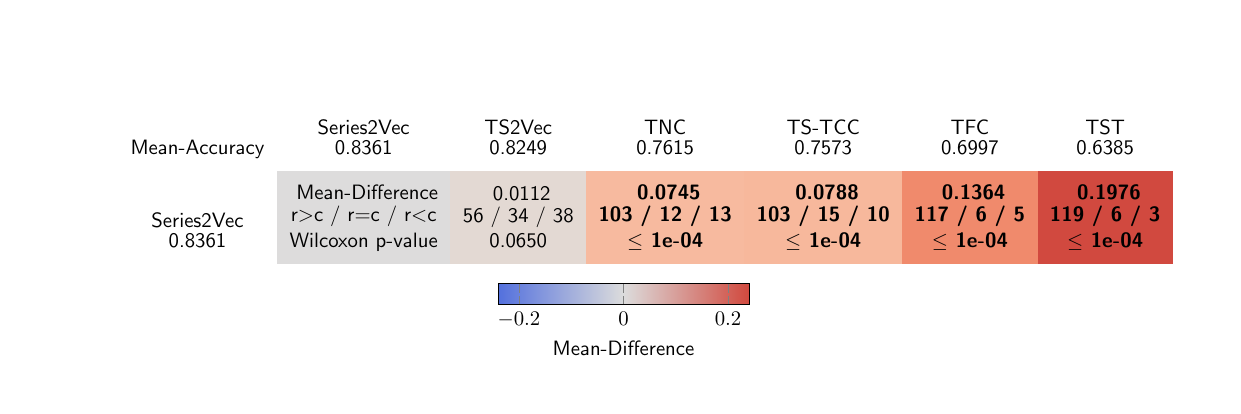}
        \caption{}
        \label{fig:mcm ucr}
    \end{subfigure}
    \begin{subfigure}{\columnwidth}
        \includegraphics[width=\columnwidth]{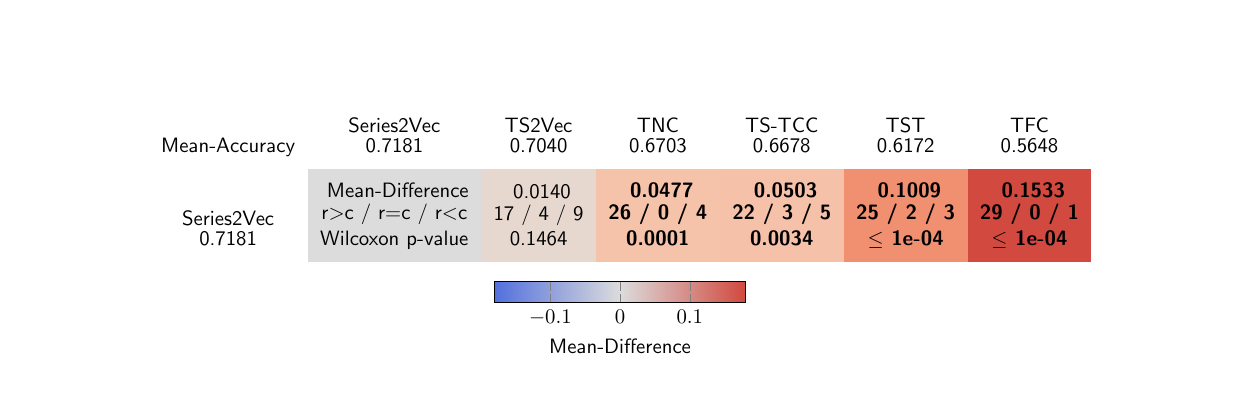}
        \caption{}
        \label{fig:mcm uea}
    \end{subfigure}
    \caption{Pairwise comparison of Series2Vec with state-of-the-art self-supervised methods. Each cell presents the average difference in accuracy across all datasets, the win/draw/loss counts of numbers of datasets for which Series2Vec obtains higher or lower accuracy and the p-value from a Wilcoxon signed rank test. The methods are ranked by their average accuracy across all the default fold of (a) 128 UCR datasets and (b) 30 UEA datasets, indicated by the values below each method. The values in bold indicate that the two methods are significantly different under significance level $\alpha=0.05$. The color represents the scale of the average difference in accuracy.}
\end{figure}

In order to highlight the great performance and generalisability of Series2Vec on diverse problems, we compare Series2Vec with the same self-supervised methods used in Section \ref{sec: baseline} on the UCR univariate and UEA multivariate time series classification benchmarking archive \cite{dau2019ucr,bagnall2018uea}.
Figures \ref{fig:mcm ucr} and \ref{fig:mcm uea} show that Series2Vec outperforms all the other methods on these archives. 
It is significantly more accurate than all the methods except TS2Vec, while winning on more datasets.

However, Series2Vec is still outperformed by the state-of-the-art time series classification methods on these archives. 
This is because the archives mainly contain relatively small-size training datasets that are less than 10,000 training examples, and are significantly smaller than the ones used in this work (see Table \ref{Tab:Data}).
Self-supervised techniques usually require large training datasets to generalise and perform well.
This highlights the limitations in current time series classification research, the need of having more larger datasets and room for improving self-supervised techniques.

\end{appendices}
\end{document}